\documentclass[letterpaper, 10 pt, conference]{ieeeconf}  
\pdfoutput=1

\IEEEoverridecommandlockouts                              

\usepackage{times} 
\usepackage{multicol}
\usepackage[bookmarks=true]{hyperref}
\usepackage{amssymb,amsmath}
\usepackage{graphicx}        
\usepackage[boxed]{algorithm2e}
\usepackage{subfigure}
\usepackage{flushend}

\DeclareGraphicsExtensions{.pdf,.jpg,.png}

\newcommand{\Prob} {\ensuremath \mathbf{P}  }

\graphicspath{{figs/}}

\title{\LARGE \bf
Curiosity Based Exploration for Learning Terrain Models
}

\author{Yogesh Girdhar, David Whitney, and Gregory Dudek
\thanks{The authors are at: Center for Intelligent Machines, McGill University, Montreal, QC H3A0E9, Canada
        {\tt\small  \{yogesh,dwhitney,dudek\}@cim.mcgill.ca}}%
}

\begin{document}

\maketitle
\thispagestyle{empty}
\pagestyle{empty}

\begin{abstract}
We present a robotic exploration technique in which the goal is to learn to a visual model and be able to distinguish between different terrains and other visual components in an unknown environment. We use ROST, a realtime online spatiotemporal topic modeling framework to model these terrains using the observations made by the robot, and then use an information theoretic path planning technique to define the exploration path. We conduct experiments with aerial view and underwater datasets with millions of observations and varying path lengths, and find that paths which are biased towards locations with high topic perplexity produce more better terrain models with high discriminative power, especially with short paths of length close to the diameter of the world. 
\end{abstract}

\section{Introduction}

This work presents an exploration technique using a realtime online topic modeling framework, which models the cause of observations made by a robot with a latent variable (called topic) that is representative of different kinds of terrains or other visual constructs in the scene, and then uses a local planner to find an exploratory path through the world which would result in learning this topic model quickly. 



We define curiosity as the unsupervised act of moving through the world in order to seek novel observations with high information content. We posit that observation data collected from such paths that seek novelty and maximize information gain would result in better terrain models. Computing information gain on low level sensor data, which in the case of vision corresponds to pixel colors or edges, might not work in many scenarios where we are interested in modeling more abstract visual constructs. Hence, we propose the use of a topic modeling framework, which have been shown to produce semantic labeling of text \cite{Blei:2003} and images \cite{Bourgault2002}, including satellite maps~\cite{Lienou2010}.




In this work we use a realtime online spatiotemporal topic modeling technique called ROST~\cite{Girdhar2013IJRR} that is suitable for use in the robotic exploration context. ROST allows for topic modeling of streaming data (observations made by a robot over time), while taking into account the spatial and temporal distribution of the data. Moreover, ROST can process the incoming observation data in real time, while providing a very close approximation to traditional batch implementations~\cite{Girdhar2013IJRR}.

At each time step, we add the observations from the  current location to the topic model, and compute the perplexity of the observations from the neighboring locations. This perplexity score, along with a repulsive potential from previously visited locations, is then used to bias the probability of next step in the path. Since observations with high perplexity have high information gain, we claim that this approach would results in faster learning of the terrain topic model, which would imply shorter exploration paths for the same accuracy in predicting terrain labels for unseen regions.

\begin{figure}[t]
\begin{center}
\includegraphics[width=0.7\columnwidth]{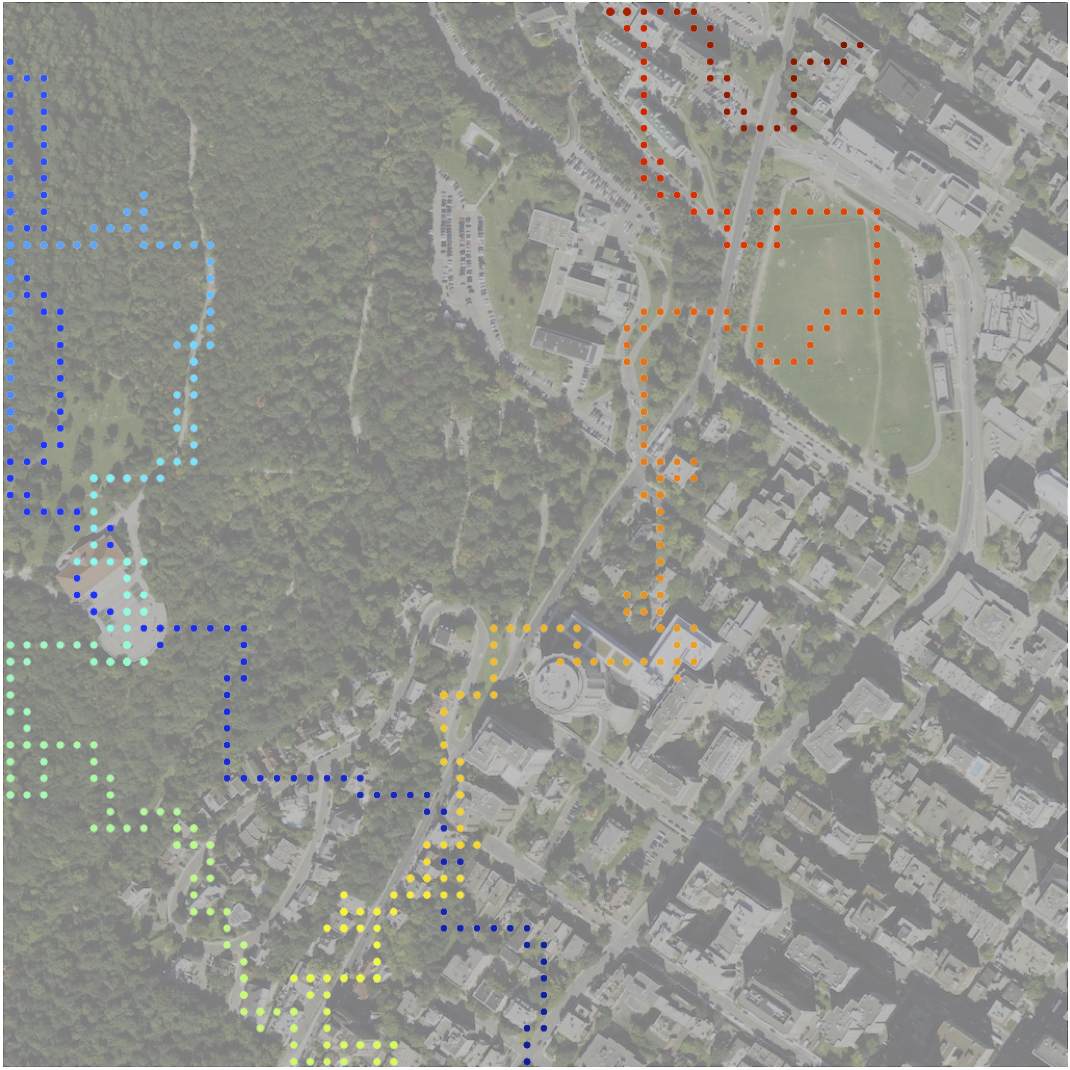}\\
\includegraphics[width=0.7\columnwidth]{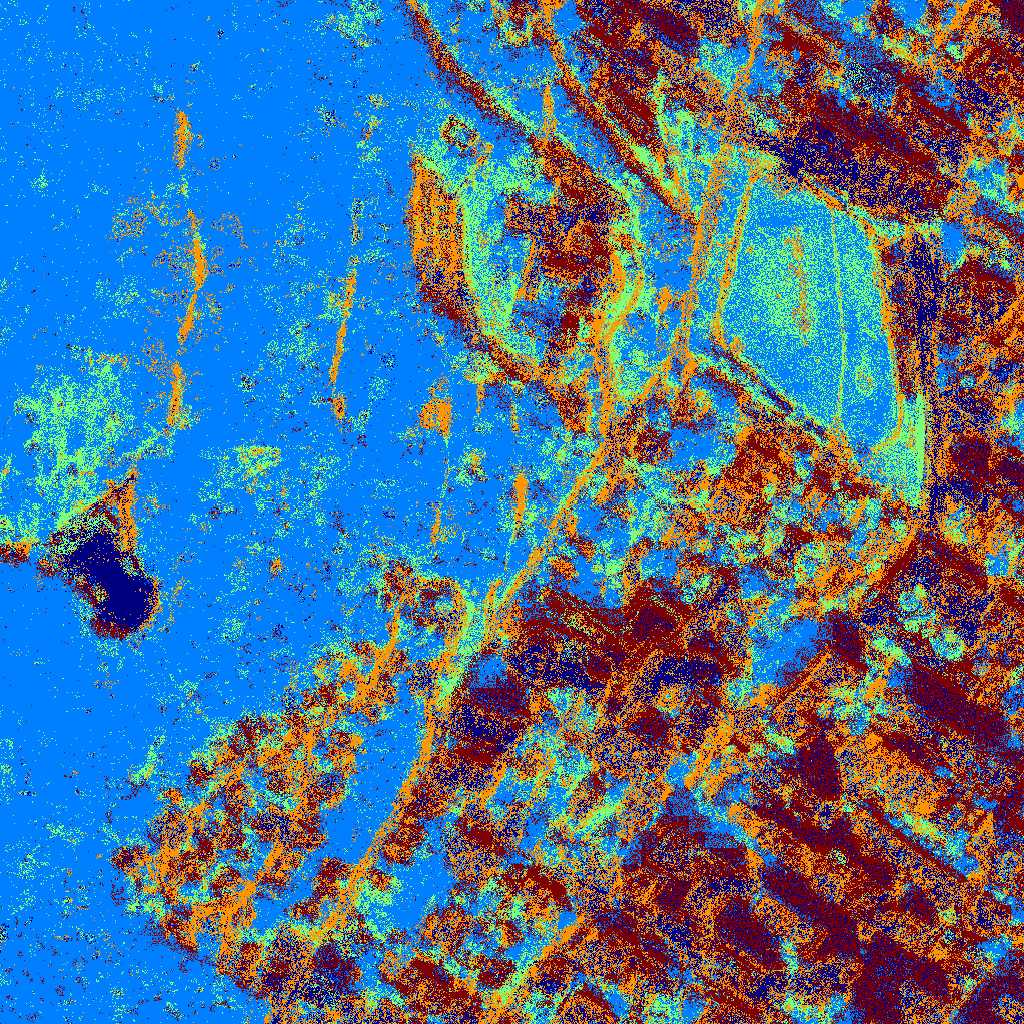}
\caption{Example of an exploratory path (top) produced by the proposed technique on a satellite map. The path begins in Blue, and ends in Red. Output of this exploration is a terrain model, which when applied to the observation from entire map produces terrain label for every location(bottom). Different colors represent different terrains. }
\label{fig:overview}
\end{center}
\end{figure}



\section{Prior Work}\label{sec:exploration}

Autonomous exploration is a well studied field in robotics, and there are different variants of the exploration problem.

\subsection{Exploration for Navigation}
Navigating a robot through free space is a fundamental problem in robotics. Yamauchi~\cite{Yamauchi} defined exploration as the ``act of moving through an unknown environment while building a map that can be used for subsequent navigation''. Yamauchi's proposed solution involved moving the robot towards the frontier regions in the map, which were described as the boundary between known free space and the uncharted territories.  

If we have an inverse sensor model of the range sensor, it is possible to compute locations in the world which would maximize the utility of the sensor reading in resolving obstacle position and shape. Grabowski~\cite{Grabowski} proposed such an exploration strategy where the goal is to maximize the understanding of obstacles rather than the exposure to free space. In this approach, the robot identifies the location with next best view in space where a sonar sensor reading would have the greatest utility in improving the quality of representation of an obstacle. 

If there is no external localizer available to the robot, then it is desirable that robot explores, maps and localizes in the environment at the same time~\cite{sim2004online}\cite{sim2009autonomous}\cite{kollar2008efficient}\cite{amigoni2010information}. Bourgault~\cite{Bourgault2002} and Stachniss~\cite{Stachniss2005} have proposed an exploration strategy which moves the robot to maximize the map information gain, while minimizing the robot's pose uncertainty.


\subsection{Exploration for Monitoring Spatiotemporal Phenomenon}
In underwater and aerial environments, obstacle avoidance is typically not the primary concern, but many different kinds of high level exploration tasks still exist. 

Binney~\cite{Binney2013} has described an exploration technique to optimize the monitoring spatiotemporal phenomena by taking advantage of the submodularity of the objective function. Bender~\cite{Bender2013} has proposed a Gaussian process based exploration technique for benthic environments, which uses an experiment specific utility function. Das et al.~\cite{Das2012} have presented techniques to autonomously observe oceanographic features in the open ocean. Hollinger et al.~\cite{Hollinger2012} have studied the problem of autonomously studying underwater ship hulls by maximizing the accuracy of sonar data stream. Smith et al.~\cite{Smith2011} have looked at computing robot trajectories which maximize information gained, while minimizing the deviation from the planned path. Girdhar et al.~\cite{Girdhar:ISER:2012} have proposed a coral reef exploration algorithm by varying the robot speed based on a surprise score.

\section{Terrain Modeling} \label{sec:sensing}
\subsection{Topic Models}
Topic modeling methods were originally developed for text analysis. Probabilistic Latent Semantic Analysis (PLSA) proposed by Hoffman~\cite{Hofmann:2001}, models the probability of observing a word $w_i$ in a given document $M$ as:

\begin{eqnarray}
\Prob(w_i=v| M) = \sum_{k=1}^{K} \Prob(w_i=v | z_i=k) \Prob(z_i=k | M).
\end{eqnarray}
where $v$ takes a value between $1\dotsc V$, the vocabulary size, and $z_i$ is the hidden variable or topic label for $w_i$. Topic label $z_i$ takes a value between $1\dotsc K$, where $K$ is a much smaller than $V$. The central idea here is the introduction of the latent variable $z$, which models the underlying topic, or the context responsible for generating the word. Latent Dirichlet Allocations \cite{Blei:2003} improve upon PLSA by placing Dirichlet priors on $\Prob(w|z)$ and $\Prob(z|M)$, which bias these distributions to be sparse, preventing overfitting. 

For modeling visual data observed by the robot, instead of text words, we use two different kinds of visual words: Oriented BRIEF (ORB) \cite{RubleeE2011} based visual words~\cite{Sivic:2006:videogoogle} that describing local visual features, and texton words \cite{Varma2005} in Lab color space to describe texture properties of a region. Moreover, instead of documents, we compute the prior topic distribution for a given word by taking into account the topic distribution in its spatial neighborhood. We posit that the resulting topic labels modeled by the system then represent high level visual patterns that are representative of different terrain types in the world.

\subsection{Generative Process for Observations}
We assume the following generative process for observations produced by the spatial region being explored. The world is decomposed into $C$ cells, in which each cell $c\in C$ is connected to its neighboring cells $G(c)~\subseteq~C$. In this paper, we only experiment with grid decomposition of the world, where each cell is connected to its four nearest neighbors. However, the general idea presented here are applicable to any topological decomposition of spacetime. 

Each cell is modeled by a mixture of at most $K$ different kinds of terrain topics, each of which when observed by a robot can result in one of the $V$ different types of visual words.  Let $\theta_{G(c)}$ be the distribution of topics in and around the cell. Intuitively, we would like visual words with the same topic to cluster together in space. This phenomena can be modeled by placing a Dirichlet prior on $\theta_{G(c)}$. 

\begin{figure}
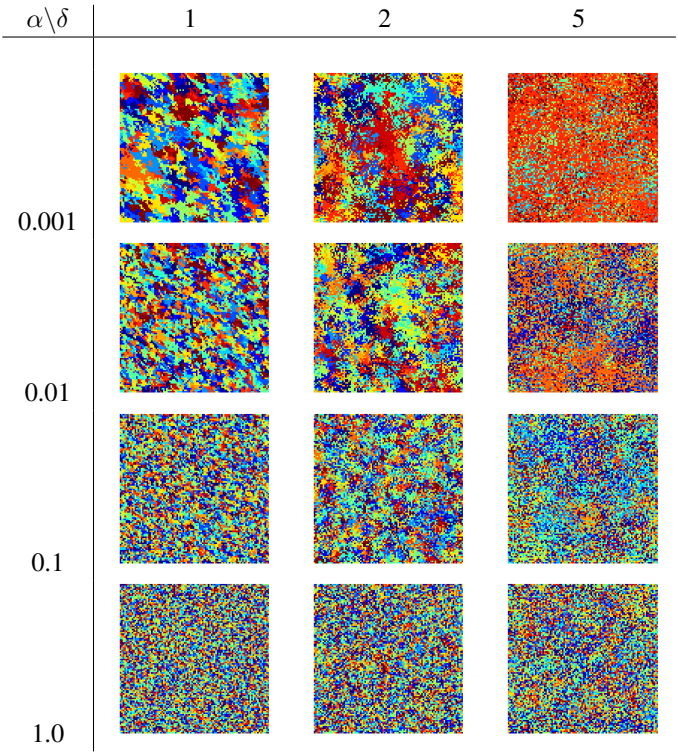

\centering

\begin{tabular}{ c| c c c }  
$\alpha \backslash \delta$ & 1 & 2 & 5 \\
\hline\\
0.001&
\includegraphics[width=0.25\columnwidth]{{{figs/dirichlet_priors_maps/worldradius50_alpha0.001_cellradius1_K16}}}&
\includegraphics[width=0.25\columnwidth]{{{figs/dirichlet_priors_maps/worldradius50_alpha0.001_cellradius2_K16}}}&
\includegraphics[width=0.25\columnwidth]{{{figs/dirichlet_priors_maps/worldradius50_alpha0.001_cellradius5_K16}}}\\  
0.01&
\includegraphics[width=0.25\columnwidth]{{{figs/dirichlet_priors_maps/worldradius50_alpha0.01_cellradius1_K16}}}&
\includegraphics[width=0.25\columnwidth]{{{figs/dirichlet_priors_maps/worldradius50_alpha0.01_cellradius2_K16}}}&
\includegraphics[width=0.25\columnwidth]{{{figs/dirichlet_priors_maps/worldradius50_alpha0.01_cellradius5_K16}}}\\  
0.1&
\includegraphics[width=0.25\columnwidth]{{{figs/dirichlet_priors_maps/worldradius50_alpha0.1_cellradius1_K16}}}&
\includegraphics[width=0.25\columnwidth]{{{figs/dirichlet_priors_maps/worldradius50_alpha0.1_cellradius2_K16}}}&
\includegraphics[width=0.25\columnwidth]{{{figs/dirichlet_priors_maps/worldradius50_alpha0.1_cellradius5_K16}}}\\  
1.0&
\includegraphics[width=0.25\columnwidth]{{{figs/dirichlet_priors_maps/worldradius50_alpha1.0_cellradius1_K16}}}&
\includegraphics[width=0.25\columnwidth]{{{figs/dirichlet_priors_maps/worldradius50_alpha1.0_cellradius2_K16}}}&
\includegraphics[width=0.25\columnwidth]{{{figs/dirichlet_priors_maps/worldradius50_alpha1.0_cellradius5_K16}}}\\  

\end{tabular}

\caption{Dirichlet Priors for 2D Maps. The table shows random maps sampled from the generative process used to characterize spatial terrain information in 2D worlds. Columns show variation in neighborhood size $\delta$, and rows show variation in Dirichlet concentration parameter $\alpha$. We see that changes in $\alpha$ control cluster sizes, whereas changes in $\delta$ control mixing of adjacent clusters. }
\label{fig:dirichlet2d}
\end{figure}

Figure \ref{fig:dirichlet2d} shows random samples from the generative process used to describe these terrain topic distribution in a 2D map. As we vary the Dirichlet concentration parameter $\alpha$ and the neighborhood size $\delta$, we see that smaller $alpha$ results in fewer topics in the neighborhood of any given cell, and smaller $\delta$ results is clusters with less mixing. 

Similar to LDA, we describe each topic $k$  by a word distribution $\phi_k$ over $V$ different types of visual words, and $\phi_k$  is assumed to have a Dirichlet prior with parameter $\beta$. This Dirichlet prior puts a constraint on the complexity of the terrain being described by this topic. Topic model $\Phi=\{\phi_k\}$ is a $K \times V$ matrix that encodes the global topic description information shared by all cells.

The overall generative process for a word $w$ in cell $c$ is thus described as following:
\begin{enumerate}
\item Word distribution for each topic $k$: $$\phi_k \sim \mathrm{Dirichlet}(\beta)$$
\item Topic distribution for words in neighborhood of $c$ : $$\theta_{G(c)} \sim \mathrm{Dirichlet}(\alpha)$$
\item Topic label for $w$: $$z \sim \mathrm{Discrete}(\theta_{G(c)})$$
\item Word label: $$w \sim \mathrm{Discrete}(\phi_{z})$$
\end{enumerate}
where $x\sim Y$ implies that random variable $x$ is sampled from distribution $Y$.

\subsection{Gibbs Sampling}
At each time step, we add the observed words to their corresponding cells, and use a Gibbs sampler to update and refine the topic labels until the next time step.

Let $P=\{p_1,\dots,p_T\}, ~p_i \in C$, be the set of cells in the current path at time $T$. 

The posterior topic distribution for a word $w_i$ in cell $p_t=c$ is given by:

\begin{eqnarray}
\begin{split}
\Prob(z_i = k |  w_i=v , p_t=c) \propto  \frac{n_{k,-i}^{v} + \beta}{\sum_{v=1}^V (n_{k,-i}^{v} + \beta)} \cdot \\
\frac{n_{G(c),-i}^{k} + \alpha}{\sum_{k=1}^K (n_{G(c),-i}^{k} + \alpha)},
\end{split} \label{eq:gibbs}
\end{eqnarray}

where $n_{k,-i}^{v}$ counts the number of words of type $v$ in topic $k$, excluding the current word $w_i$, and $n_{G(c),-i}^{k}$ is the number of words with topic label $k$ in neighborhood of cell $c$, excluding the current word $w_i$, and $\alpha, \beta$ are the Dirichlet hyper-parameters.  Note that for a neighborhood size of 0, $G(c) = c$, and the above Gibbs sampler is equivalent to the LDA Gibbs sampler proposed by Griffiths et al.~\cite{Griffiths:2004}.

Several different strategies exist in the literature to do online refinement of the topic label assignment on streaming data \cite{Canini2009}. However, in this work, we are interested in the more constrained realtime version of the problem. After each new observation, we only have a constant amount of time to do topic label refinement, hence any online refinement algorithm that has computational complexity which increases with new data is not applicable. 

We then must use a refinement strategy which only partially updates the topic labels after each time step. To ensure that the topic labels from the last observation converge before the next observation arrives, at each time step, for each refine iteration, we refine the last observation with probability $\tau$, or a previous observation with probability $(1-\tau)$. We pick the previous observation using age proportional random sampling. We found $\tau=0.5$ to work well in most cases, however on faster machines, $\tau$ could be set to a lower value, which would encourage better globally optimal topic labels. Algorithm \ref{alg:refinetopics} summarizes the proposed realtime topic refinement strategy.

\begin{algorithm}[]	
	\dontprintsemicolon
	\caption{Keep topic labels up-to-date as new observations arrive.} 
	\label{alg:refinetopics}
	\While{true}{
	\While{no new observation}{
		$a \sim \mathrm{Bernoulli}(\tau)$\;
		\eIf{ $a == 0$}{
		  	\emph{\small{(*select last observation*) }}\; 
			$t \leftarrow T$\;
		}{
		  	\emph{\small{(*pick an observation with probability proportional to its timestamp*) }}\; 
			$t \leftarrow q , \Prob(q=j) \propto j, 1 \leq j \leq T   $\;
		}
	  		\ForEach{ word $w_i~ \mathrm{in}~ p_t $ }{   
	  			\emph{\small{(*update the topic label for word in the observation *) }}\; 
	  			$z_i \sim \Prob(z_i = k |  w_i=v, p_t=c)$ \;
	  		}
	
	}
	$T \leftarrow T+1$\;
    Add new observed words to their corresponding cells.
	}
\end{algorithm}

\section{Curiosity based Exploration} \label{sec:curiosity}


At time $t$, let the robot be in cell $p_t=c$, and let $G(c)=\{g_i\}$ be the set of cells in its neighborhood. We would like to compute a weight value for each $g_i$, such that 
\begin{eqnarray}
\Prob(p_{t+1} = g_i) \propto \mathrm{weight}(g_i)
\end{eqnarray}

In this work we consider a four different weight functions.
\begin{enumerate}
\item Random Walk - Each cell in the neighborhood is equally likely to be the next step:
    \begin{eqnarray}
    \mathrm{weight}(g_i) = 1.
    \end{eqnarray}
\item Stochastic Coverage - Use a potential function to repel previously visited locations:
    \begin{eqnarray}
    \mathrm{weight}(g_i) = \frac{1}{\sum_j n_j/d^2(p_t,c_j)}.
    \end{eqnarray}
    where $n_j$ is the number of times we have visited cell $c_j$, and $d(p_t,c_j)$ is the Euclidean distance between these two cells. 

\item Word Perplexity - Bias the next step towards cells which has high word perplexity:
    \begin{eqnarray}
    \mathrm{weight}(g_i) = \frac{\mathrm{WordPerplexity}(g_i)}{\sum_j n_j/d^2(p_t,g_j)}.
    \end{eqnarray}

\item Topic Perplexity - Bias the next step towards cells which has high topic perplexity:
    \begin{eqnarray}
    \mathrm{weight}(g_i) = \frac{\mathrm{TopicPerplexity}(g_i)}{\sum_j n_j/d^2(p_t,g_j)}.
    \end{eqnarray}
    
\end{enumerate}

We compute the word perplexity of the words observed in $g_i$ by taking the inverse geometric mean of the probability of observing the words in the cell, given the current topic model and the topic distribution of the path thus far. 
\begin{eqnarray}
\begin{split}
\mathrm{WordPerplexity}(g_i) =\\
\exp\left(-\frac{\sum_i^{W} \log  \sum_k \Prob(w_i=v | k)\Prob(k | P)  }{W} \right),
\end{split}
\end{eqnarray}

where $W$ is the number of words observed in $g_i$, $\Prob(w_i=v|k)$ is the probability of observing word $v$ if its topic label is $k$, and $\Prob(k|P)$ is the probability of seeing topic label $k$ in the path executed by the robot thus far.

To compute topic perplexity of the words observed in $g_i$, we first compute topic labels $z_i$ for these observed words by sampling them from the distribution in Eq. \ref{eq:gibbs}, without adding these words to the topic model. These temporary topic labels are then used to compute the perplexity of $g_i$ in topic space.

\begin{eqnarray}
\begin{split}
\mathrm{TopicPerplexity}(g_i) =\\
 \exp\left(-\frac{\sum_i^{W} \log \Prob(z_i=k | P)  }{W} \right).
\end{split}
\end{eqnarray}

\begin{figure*}[t]
\begin{center}

\subfigure[]{\includegraphics[width=0.55\columnwidth]{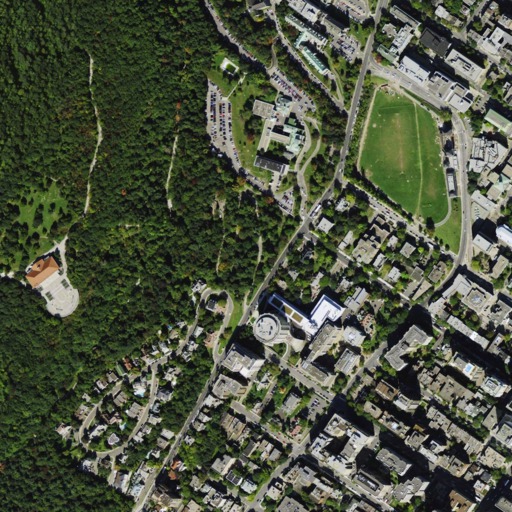}}
\subfigure[]{\includegraphics[width=0.55\columnwidth]{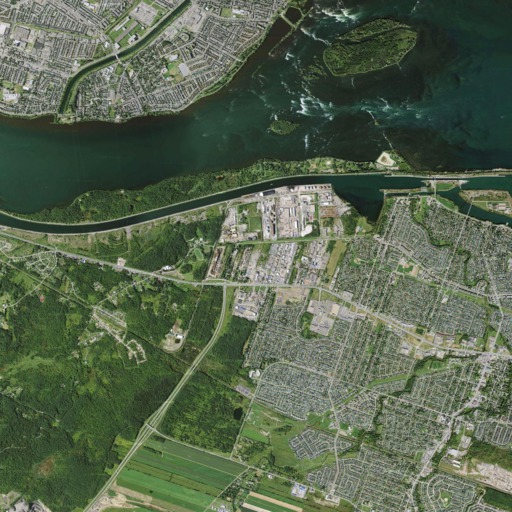}}
\subfigure[]{\includegraphics[width=0.55\columnwidth]{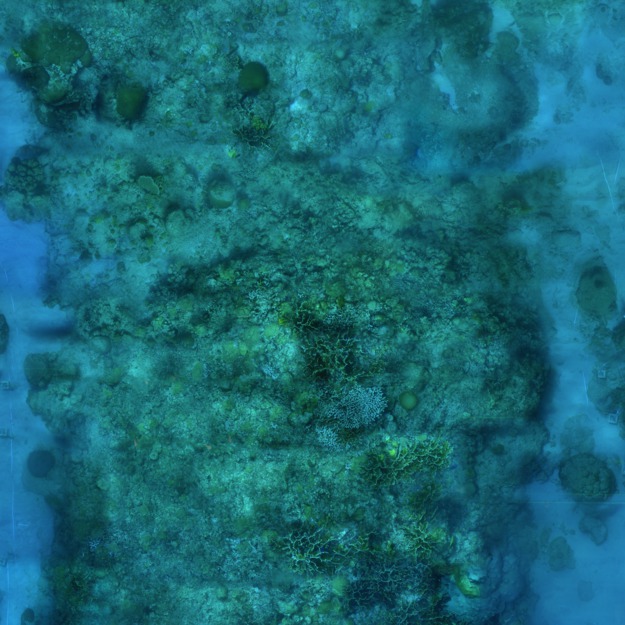}}

\subfigure[]{\includegraphics[width=0.55\columnwidth]{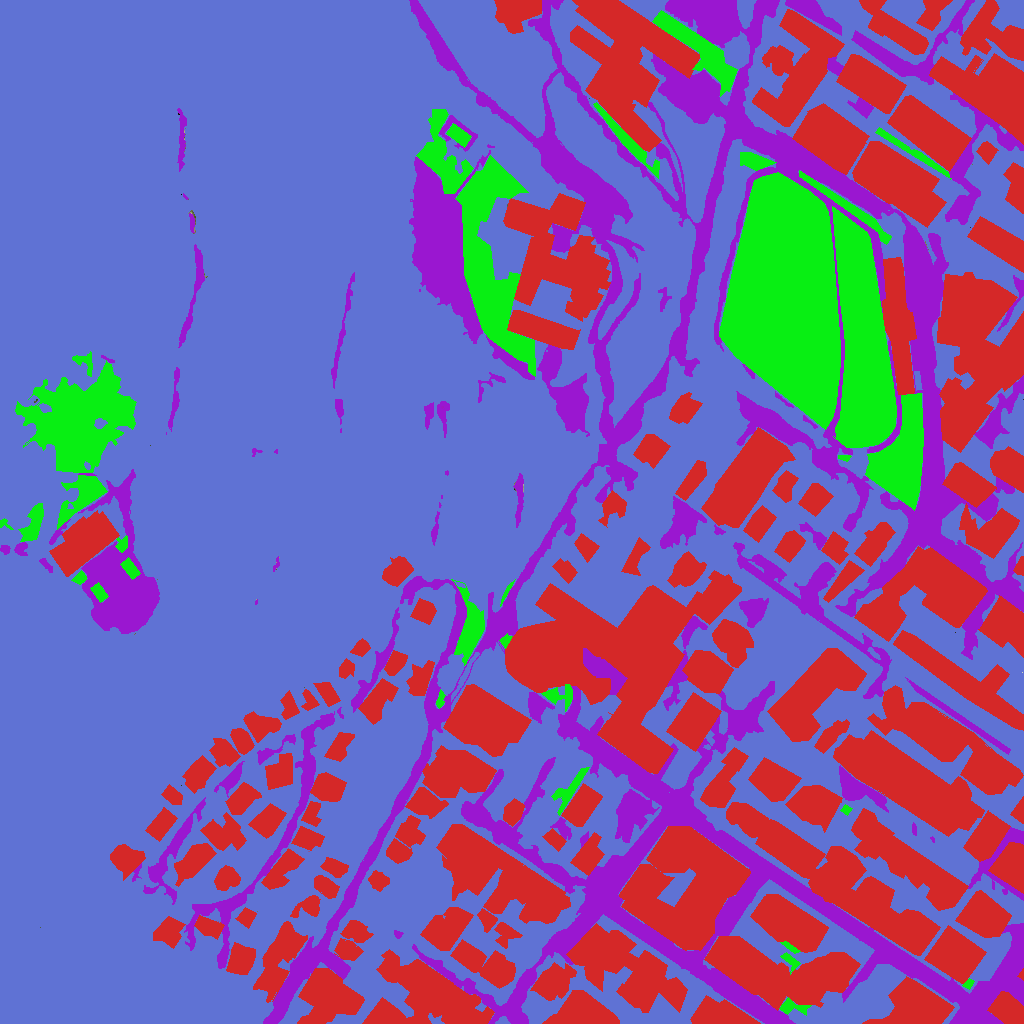}}
\subfigure[]{\includegraphics[width=0.55\columnwidth]{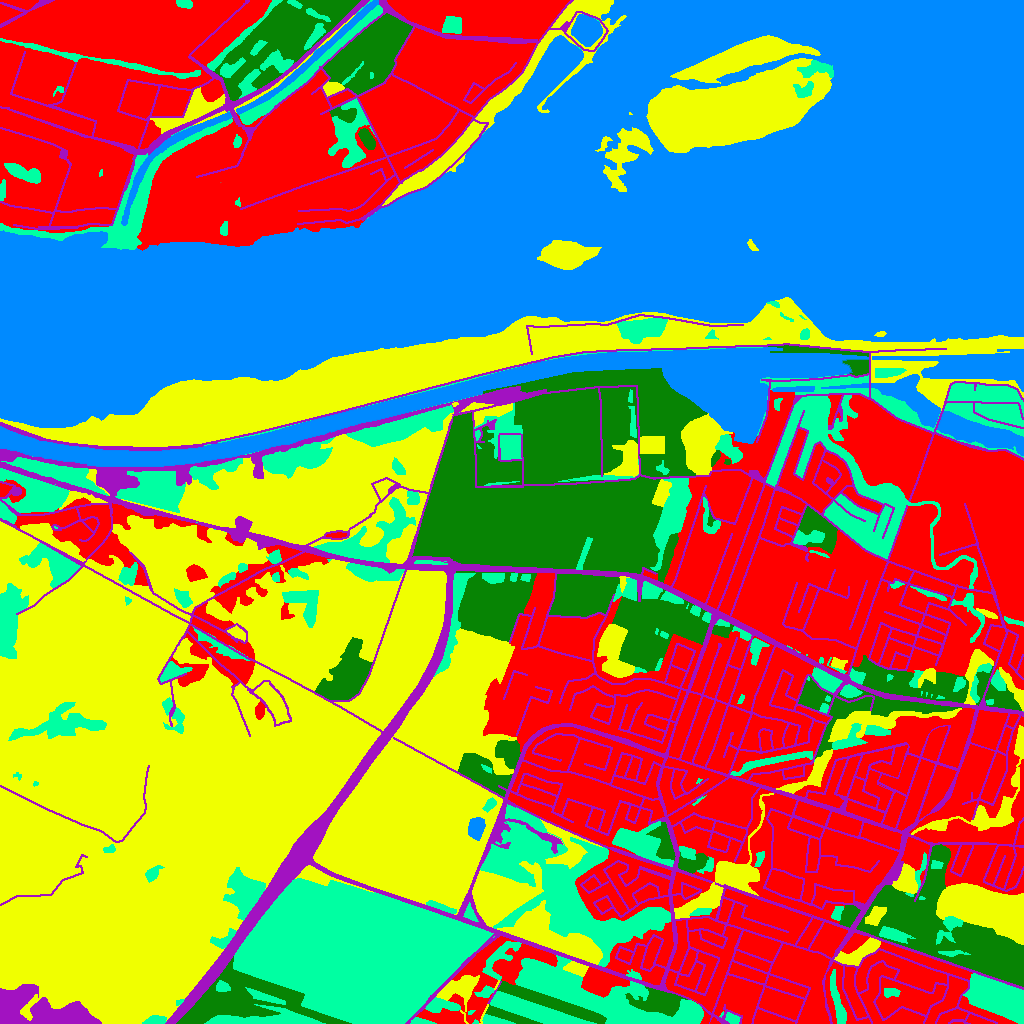}}
\subfigure[]{\includegraphics[width=0.55\columnwidth]{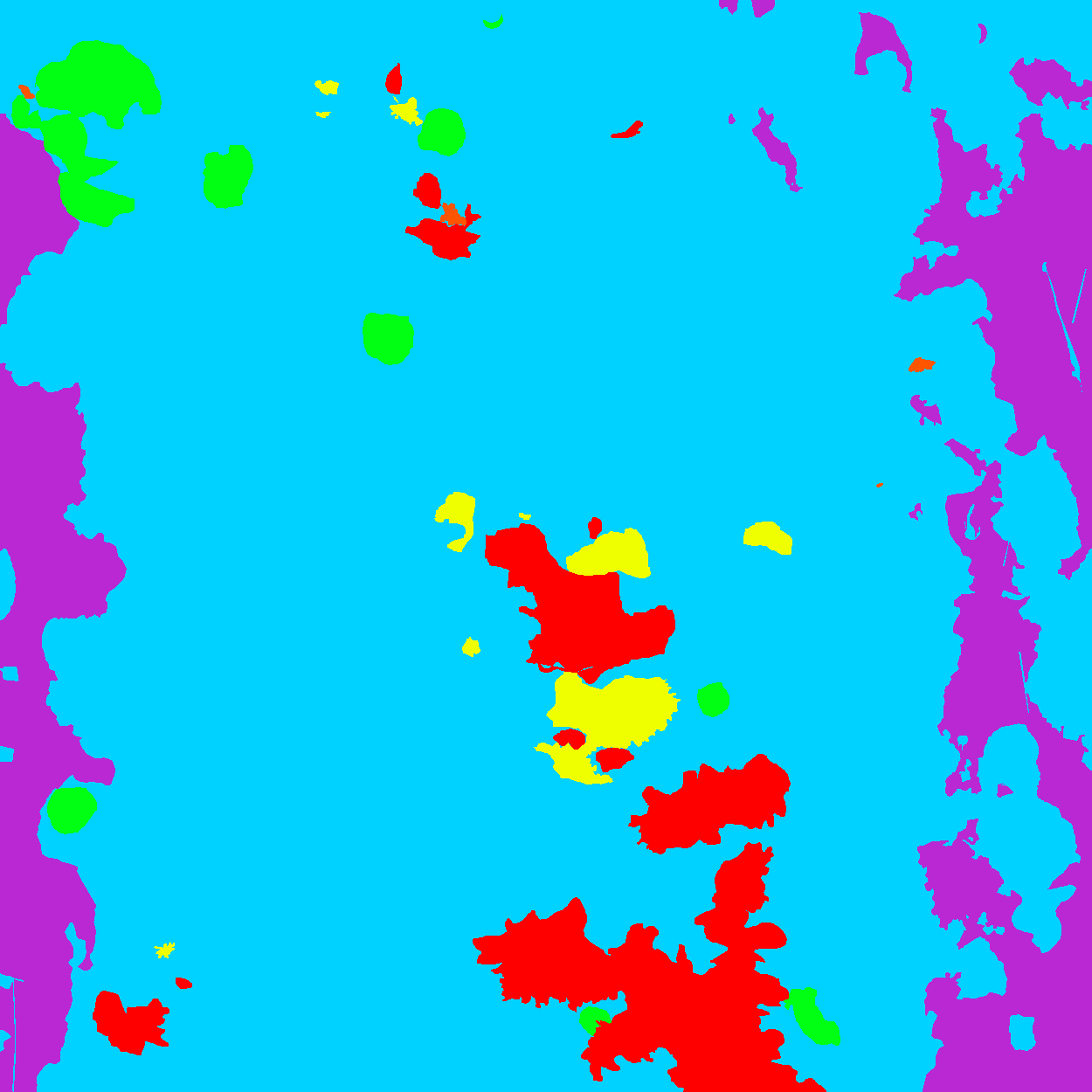}}

\subfigure[]{\includegraphics[width=0.55\columnwidth]{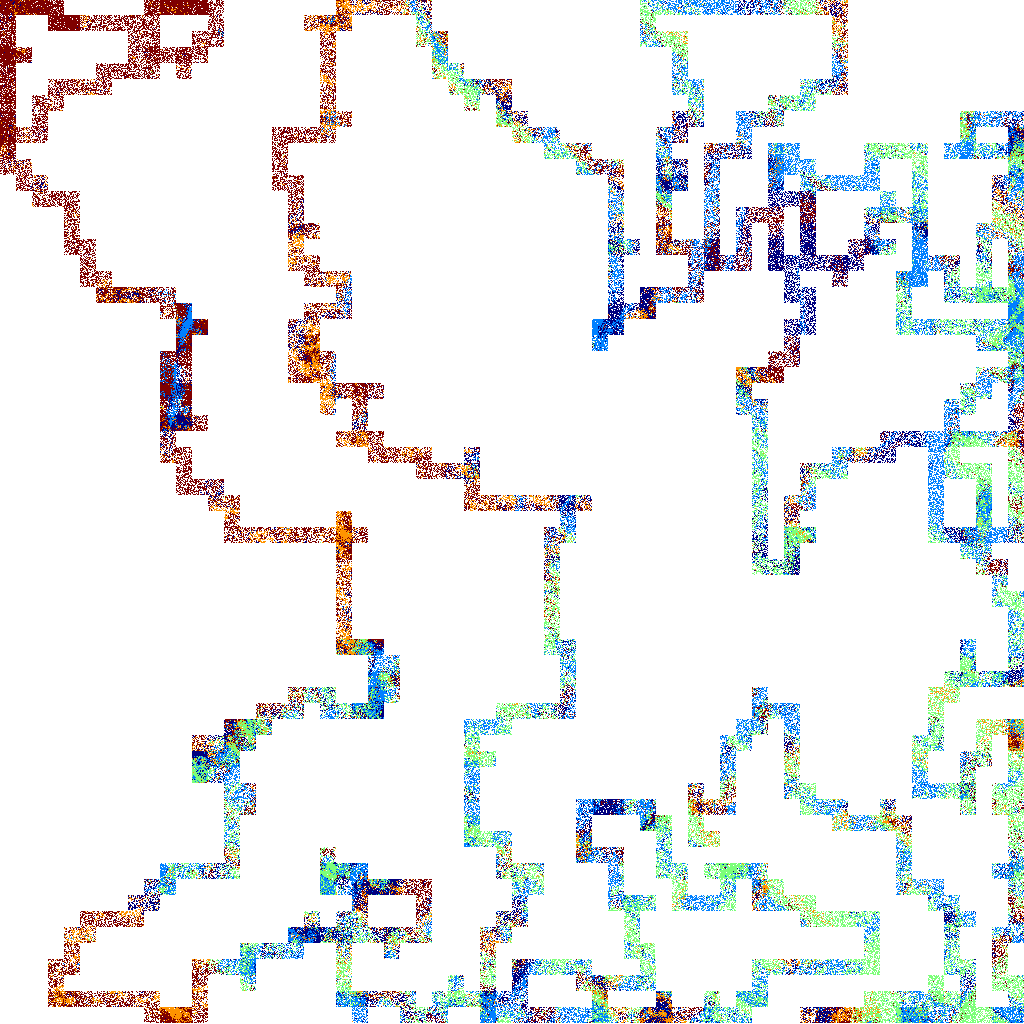}}
\subfigure[]{\includegraphics[width=0.55\columnwidth]{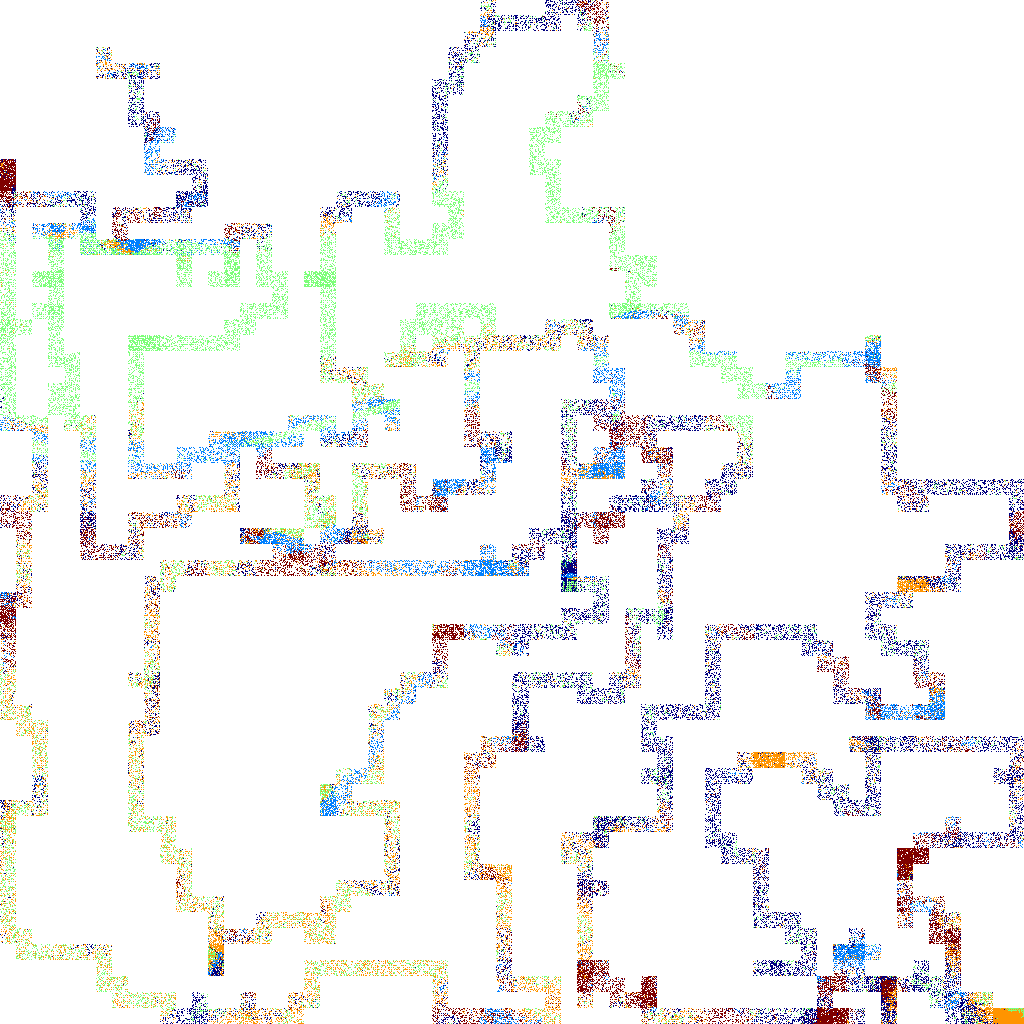}}
\subfigure[]{\includegraphics[width=0.55\columnwidth]{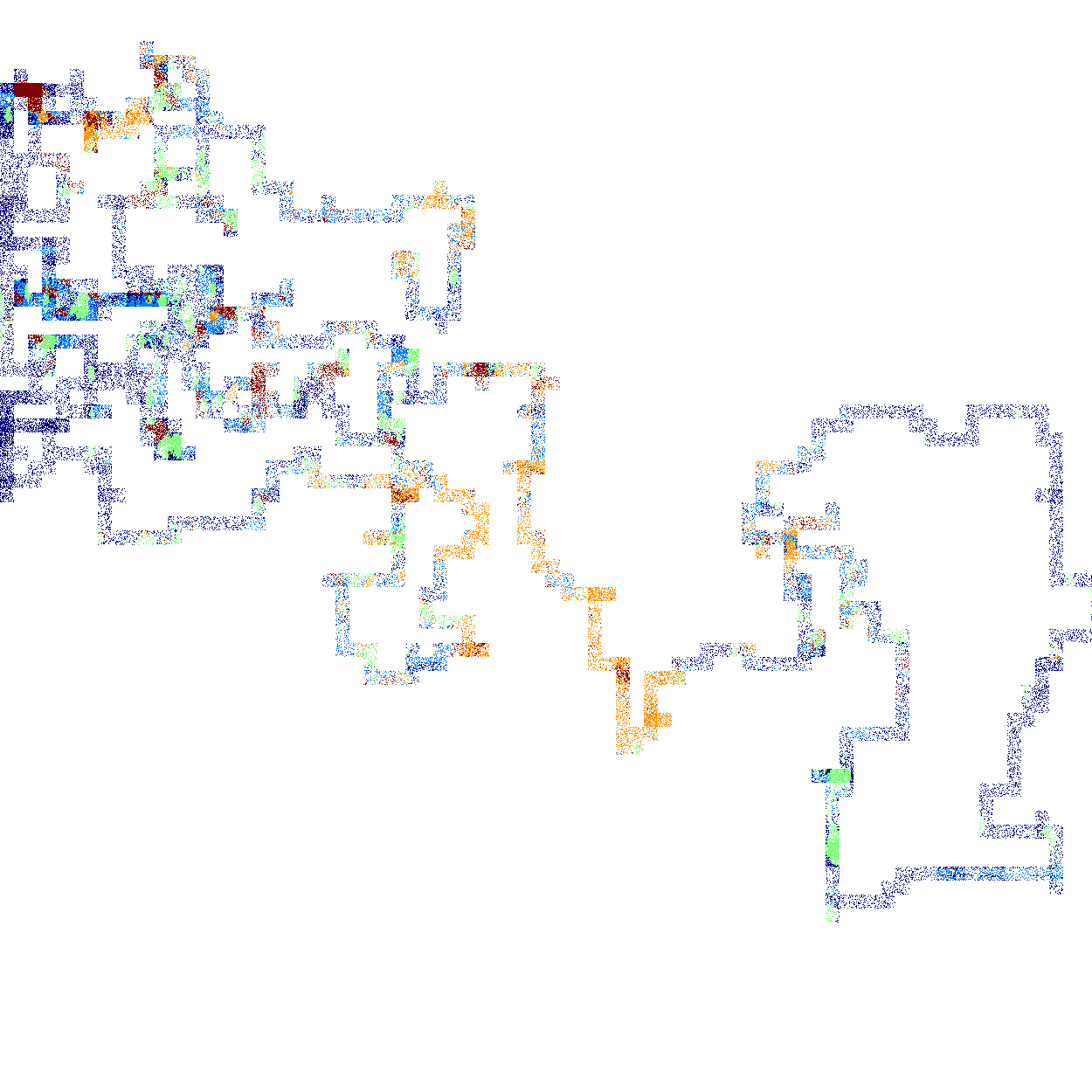}}

\subfigure[]{\includegraphics[width=0.55\columnwidth]{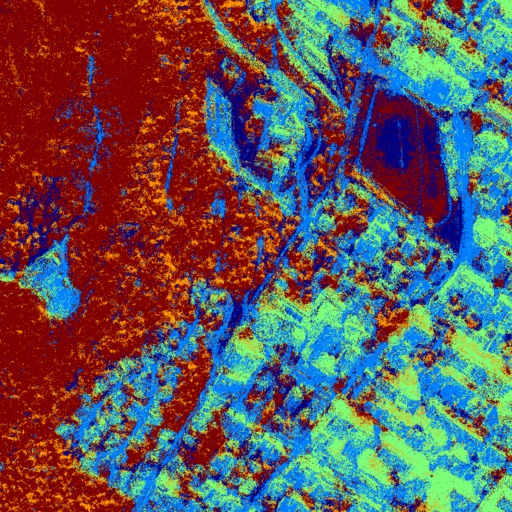}}
\subfigure[]{\includegraphics[width=0.55\columnwidth]{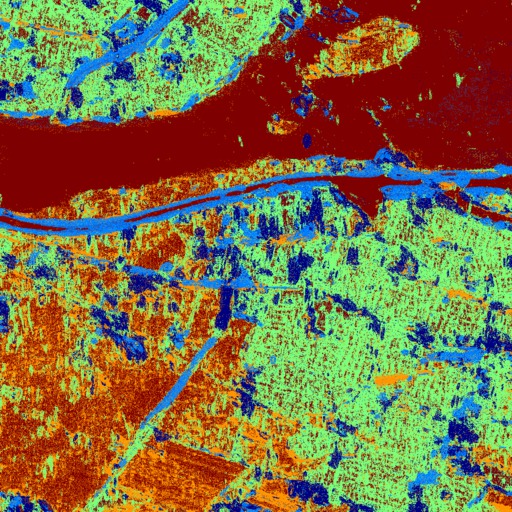}}
\subfigure[]{\includegraphics[width=0.55\columnwidth]{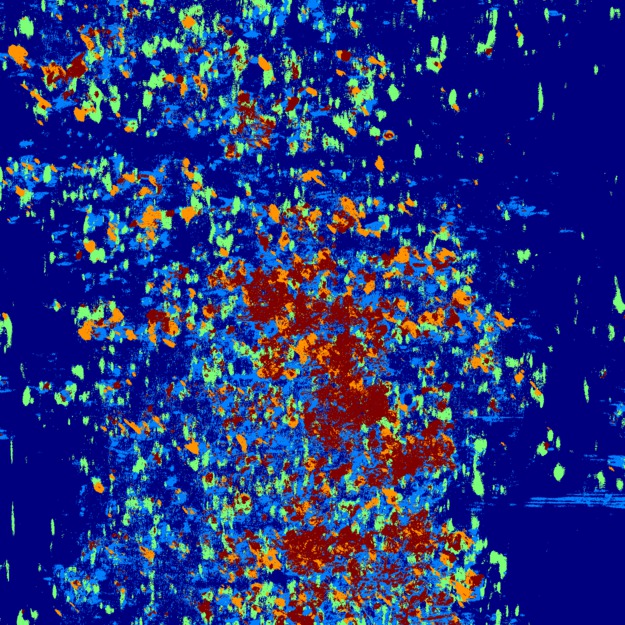}}

\caption{(a)-(c) Input image used to generate observation data, (d)-(f) Groundtruth labeling. (g)-(i) An example path and the topic labels computed online. Parts of the path with higher density of points is indicative of multiple passes through that cell. (j)-(l) Terrain labeling of the map using the topic model computed on the path. }\label{fig:ds}
\end{center}
\end{figure*}

\begin{figure*}[t]
\begin{center}

\includegraphics[width=1.9\columnwidth]{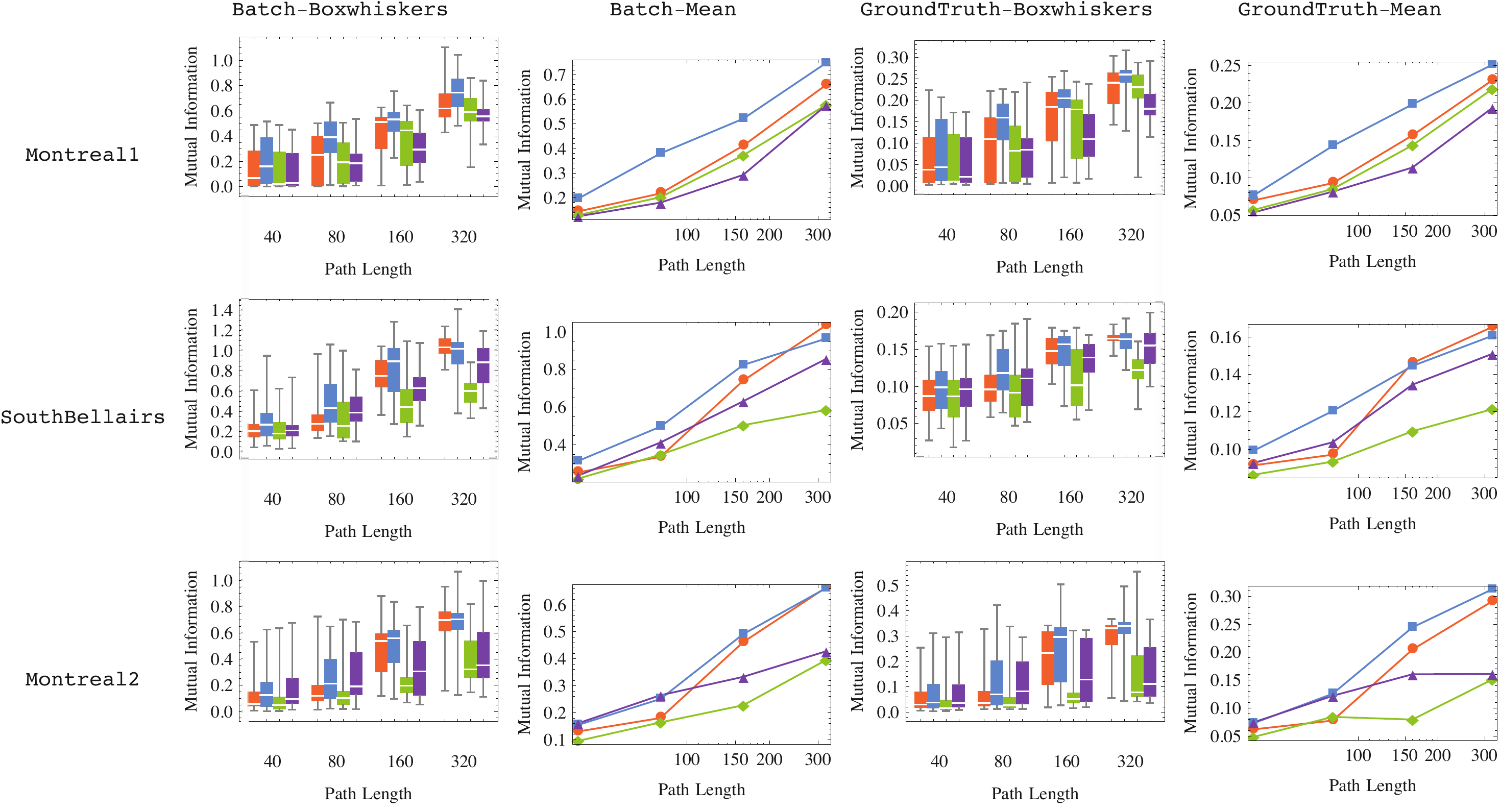} \\
\includegraphics[width=0.25\columnwidth]{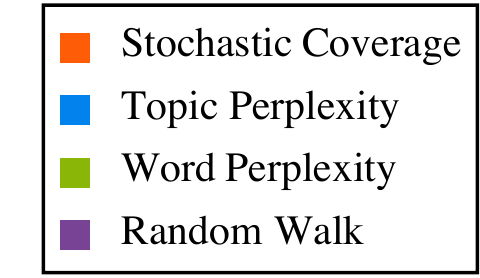} 
\caption{Evaluation of the proposed exploration techniques. The plots show mutual information between the maps labeling produced using the topic model computed online during the exploration, with maps labeled by a human, and maps labeled by batch processing of the data.}\label{fig:plots}
\end{center}
\end{figure*}

\section{Experiments}
To validate our hypothesis that biasing exploration towards high perplexity cells will result in a better terrain topic model of the environment, we conducted the following experiment. We considered three different maps, two aerial view, and one underwater coral reef map.  
\begin{table}
\begin{center}
\begin{tabular}{|l|l|l|l|l|}
\hline
Dataset & width(px) & height(px) & n.cells & n.words\\
\hline
Montreal1 & 1024 & 1024 & 4096 & 3,239,631\\
Montreal2 & 1024 & 1024 & 4096 & 1,675,171\\
SouthBellairs & 2500 & 2500 & 6241 & 1,664,749\\
\hline
\end{tabular}
\caption{Dataset specifications} \label{tab:ds}
\end{center}
\end{table}

We extracted ORB words describing local features, and texton words describing texture at every pixel (every second pixel for the SouthBellairs underwater dataset). ORB words had a dictionary size of 5000, and texton words had a dictionary size of 1000. The dictionary was computed by extracting features from every 30th frame of a completely unrelated movie\footnote{We used the documentary movie Baraka(1992) for extracting visual feature, because of its rich visuals from many different contexts}.

Each of these maps were decomposed into square cells of width 16 (32 for SouthBellairs). Now for each weight function, we computed exploration paths of varying length, with 20 different random restart locations for each case. Each time step was fixed at 200 milliseconds to allow the topic model to converge. We limited the path length to 320 steps, which is about $5\sqrt|C|$.

Each of these exploration runs returned a topic model $\Phi_p$, which we then used to compute topic labels for each pixel in the map in batch mode. Let $Z_p$ be these topic labels. An example of this labeling for each of the three dataset is shown in Fig. \ref{fig:ds} (j,k,l). We compared this topic labeling with two other labelings: human labeled ground-truth $Z_h$, and labels computed in batch mode $Z_b$, with random access to the entire map. 


We then computed the mutual information between $Z_p$ and $Z_h$, 
$Z_p$ and $Z_b$, and plotted the results as a function of path length, as shown in Fig. \ref{fig:plots}.

\section{Results and Discussion}\label{sec:results}
The results are both encouraging and surprising. As shown in Fig. \ref{fig:plots}, we see that topic perplexity based exploration (shown with blue squares) performs consistently better than all other weight functions, when compared against ground truth, or the batch results. 

For paths of length 80, which is close to the width of the maps, we see that mutual information between  topic perplexity based exploration and ground truth is 1.51, 1.20 and 1.05  times higher respectively for the three datasets, compared to the next best performing technique. 

For long path lengths (320 steps or more), stochastic coverage (shown with orange circles) based exploration matches the mean performance of topic perplexity exploration. This is expected because the maps are bounded, and as the path length increases, the stochastic coverage algorithm is able to stumble across different terrains, even without a guiding function. 

For short path lengths (40 steps or less), we do not see any statistical difference between the performance of different techniques.

Marked with purple triangles, we see the results of exploration using Brownian random motion.  Although this strategy has a probabilistic guarantee of asymptotically complete coverage, but it does so at a lower rate that stochastic coverage exploration startegy.
A random walk in two dimensions is expected to travel a distance of $\sqrt n$ from start, where $n$ is the number of steps. Hence it is highly likely that it never visits different terrains. The resulting topic models from these paths are hence unable to resolve between these unseen terrains. 

The performance of word perplexity exploration (shown with green diamonds) is surprisingly poor in most cases. We hypothesis that this poor performance is due to the algorithm getting pulled towards locations with terrain described by a more complex word distribution. This will cause the algorithm to stay in these complex terrains, and not explore as much as the other algorithms. In comparison, the topic perplexity exploration is not affected by the complexity of the distribution describing the topic, and is only attracted to topic rarity.

\section{Conclusion} \label{sec:conclusion}
We have presented a novel exploration technique that aims to learn a terrain model for the world by finding paths with high information content. The use of a realtime online topic modeling framework allows us to model incoming streams of low level observation data via the use of a latent variable representing the terrain. Given this online model, we measure the utility of the potential next steps in the path. We validated the effectiveness of the proposed exploration technique over candidate techniques by computing mutual information between the terrain maps generated through the use of the learned terrain model, and hand labeled ground truth, on three different datasets.

\section*{Acknowledgment}
This work was supported by the Natural Sciences and Engineering Research Council (NSERC) through the NSERC Canadian Field Robotics Network (NCFRN).

\bibliographystyle{IEEEtran}
\bibliography{IEEEabrv,library}
\end{document}